\documentclass{article} 
\usepackage{nips13submit_e,times}
\usepackage{hyperref}
\usepackage{epsfig}
\usepackage{url}
\usepackage{enumerate}
\usepackage{amsmath}
\usepackage{amsfonts}
\usepackage{tikz}
\usepackage{pgfplots}

\usepackage[normalem]{ulem}
\usepackage{comment}

\usepackage{booktabs} 
\usepackage{url}
\usepackage[linesnumbered,ruled]{algorithm2e}

\title{Neuroevolutionary algorithms driven by neuron coverage metrics for semi-supervised classification}

\author{Roberto Santana \And Ivan Hidalgo-Cenalmor  \And Unai Garciarena 
  \AND Alexander Mendiburu \hspace{2.0cm} Jose Antonio Lozano \\\\
  
  Intelligent Systems Group \\  
  University of the Basque Country (UPV/EHU)\\
  San Sebastian, Spain \\\\
  \texttt{\{roberto.santana,unai.garciarena,alexander.mendiburu,ja.lozano\}@ehu.eus}
}

\nipsfinalcopy 

\begin{document}

\maketitle

\begin{abstract}
  In some machine learning applications the availability of labeled instances for supervised classification is limited while unlabeled instances are abundant. Semi-supervised learning algorithms deal with these scenarios and attempt to exploit the information contained in the unlabeled examples. In this paper, we address the question of how to evolve neural networks for semi-supervised problems. We introduce neuroevolutionary approaches that exploit unlabeled instances by using neuron coverage metrics computed on the neural network architecture encoded by each candidate solution. Neuron coverage metrics resemble code coverage metrics used to test software, but are oriented to quantify how the different neural network components are covered by test instances. In our neuroevolutionary approach, we define fitness functions that combine classification accuracy computed on labeled examples and neuron coverage metrics evaluated using unlabeled examples. We assess the impact of these functions on  semi-supervised problems with a varying amount of labeled instances. Our results show that the use of neuron coverage metrics helps neuroevolution to become less sensitive to the scarcity of labeled data, and can lead in some cases to a more robust generalization of the learned classifiers.\\
  {\bf{keywords}}: neuroevolution, neuron coverage, semi-supervised learning, neural networks, deep learning, NAS
\end{abstract}

%
%

\section{Introduction}

Neuroevolutionary approaches \cite{Galvan_and_Mooney:2021,Stanley_et_al:2019} are extensively used to optimize the neural network architecture in supervised machine learning tasks. Usually, a training dataset is used to guide the search for the optimal model. However, in some domains, the availability of labeled examples is limited due to the cost of human labeling or the scarcity of data.  In these scenarios, unlabeled data become more precious. Semi-supervised machine learning algorithms have been proposed to take advantage of the unlabeled examples during the learning stage of the model. Among the semi-supervised machine learning approaches, we can find disagreement-based methods such as co-training \cite{Blum_and_Mitchell:1998}, graph-based methods like those based on min-cuts \cite{Blum_and_Chawla:2001}, and low-density separation methods such as semi-supervised support vector machines \cite{Bennett_and_Demiriz:1998}. 


Evolutionary algorithms have been also applied to semi-supervised problems. In \cite{Fitzgerald_et_al:2015}, the application of  grammatical evolution (GE)  for semi-supervised classification is proposed. The quality of the model is evaluated by combining the accuracy on the labeled data with one measure of cluster quality on the unlabeled examples. Kamalov et al.  \cite{Kamalov_et_al:2021}  apply a GP algorithm combined with a self-labeling algorithm to classify time-series. They show that the combination of both types of algorithms improves state-of-the-art semi-supervised methods for the addressed classification problems.  Silva et al. \cite{Silva_et_al:2018} propose a way to use unlabeled data as part of GP learning to increase the accuracy of GP classifiers for problems with noisy labels.

In this paper, we focus on neuroevolutionary approaches to semi-supervised learning. More specifically, we investigate the behavior of evolutionary approaches that evolve neural networks for semi-supervised classification.  There are some works in the field of neural architecture search (NAS) that consider semi-supervised tasks \cite{Liu_et_al:2020c,Pauletto_et_al:2022}. The approach in these papers consists of the identification of an unsupervised \emph{pretext task} from which a neural architecture is learned that is then transferred to the target supervised task for which labeled data is available. For example, for an image segmentation task, an input image is first rotated in four preset directions and then a neural network architecture is searched for the \emph{pretext task} of predicting the rotation. Subsequently, the selected architecture is retrained to solve the original image segmentation task. In this paper, we follow a completely different approach, in which unlabeled data is used to evaluate different metrics describing the neuron coverage of the neural network architecture encoded by the candidate solution.

Neuron coverage metrics  \cite{Ma_et_al:2018,Pei_et_al:2017,Yang_et_al:2022} are inspired by code coverage metrics which serve the  functional and structural testing of code systems. Usually, a software coverage metric serves to quantify to what extent different components of a code (e.g., lines, program branches, modules, etc.) are ``covered'' by a set of test examples. For neural networks, the components whose coverage is targeted by the corresponding metrics are the neurons, layers, or even more fine-grain functional characterization of the network, such as the range of possible values for the activation functions of the neurons. Neuron coverage metrics have mainly been applied as a way to verify neural networks and detect possible errors  \cite{Lee_et_al:2022,Gerasimou_et_al:2020}. The rationale behind their application is that a test set that covers all possible components  of a neural network serves as a detailed characterization of the network behavior.

In this paper, we propose to use neuron coverage metrics as a supervisory signal to discriminate among evolved architectures. The assumption is that these metrics, when computed using unlabeled data, can serve to predict whether the neural network architecture is likely to, at least, ``evaluate properly'' similar data. Therefore, neuroevolution will be driven by the performance of the evolved network architectures on labeled data and their potential to be covered by unlabeled data.

The paper is organized as follows:

In the next section, we present the necessary background on semi-supervised classification and neuron coverage metrics.  Related work is analyzed in Section~\ref{sec:RELATED_WORK}. Section~\ref{sec:NE_APPROACH} introduces the neuroevolutionary approach, explaining the evolutionary operators and the characteristics of the tensorflow-based implementation. The characteristics of the training process and the fitness function specifically conceived for the semi-supervised scenario are discussed in Section~\ref{sec:NE_FITNESS_FUNCTION}.  Section~\ref{sec:EXPE} explains the experimental framework and discusses the results of the experiments. Finally, Section~\ref{sec:CONCLU} concludes the paper and discusses future work.

 \section{Semi-supervised classification and neuron coverage metrics} \label{sec:BACKGROUND}

 \subsection{Semi-supervised classification}

  We address the semi-supervised classification task of learning a function $f:\mathcal{X} \mapsto \mathcal{Y}$ from a training data set $D_{train} =$  $\{(x^1,y^1), \dots,$ $(x^i,y^i), \dots,$  $(x^l,y^l), x^{l+1}, \dots,x^m \}$, where $\mathcal{X}$ is the feature space,  $\mathcal{Y} \in \{0,1\}$, $l$  is the number of labeled training examples, and $u=m-l$ is the number of unlabeled instances. This type of problem can be considered as an example of incomplete supervision  since only a subset of training data is given with labels \cite{Hernandez_et_al:2016}.

 \subsection{Neuron coverage metrics}  \label{sec:COV_METRICS}

 In the literature, there are slightly different definitions of the neuron coverage metrics. We have mainly adopted the conventions used in \cite{Ma_et_al:2018,Yang_et_al:2022} with some few changes. 
 
 Let $c$ represent a neuron of a multi-layer perceptron (MLP) of $A$ layers.  $M_1, M_2, \dots, M_A$ represent the number of neurons in each layer and  $N = \sum_{j=1}^{A} M_j$ is the total number of neurons in the network.
 
 We use $\phi(x^i,c)$ to denote the  function that returns the output of neuron $c$ given $x^i$ as input. For a given neuron $c$,  it is said to be \emph{activated} for a given input $x^i$ if  $\phi(x^i,c)>t$,  where $t$ is a given threshold. $L_{c}$ and $H_{c}$ will respectively represent the lower and upper bounds of function  $\phi(x^i,c)$ for $x^i \in D$. These values are determined by analyzing the values of $\phi(x^i,c)$ for the training dataset $D$. Usually, the set $D$ corresponds to a set of instances in the training dataset, i.e., $D=D_{train}$.

  \subsubsection{Neuron coverage}

  Given a set of instances $D$ and a given threshold $t$, the neural network coverage  \cite{Pei_et_al:2017} measures the proportion of neurons in MLP that have been activated by at least one instance in $D$:

   \begin{equation}
    NC = \frac{ \left| \{c| \,  \exists x^i \in D: \phi(x^i,c) > t \} \right|} {N}
   \end{equation}

 \subsubsection{Top-$K$ neuron coverage}
For a given test input $x^i$ and neurons $c$   and $c'$ in the same layer, $c$ is more active than $c'$ if  $\phi(x^i,c) > \phi(x^i,c')$.  For the $j$-th layer, $top^j_K(x^i)$ on layer $j$  denotes the set of neurons that have the largest $K$ outputs on that layer given $x^i$.

The top-$K$ neuron coverage (TKNC) measures how many neurons have once been among the most active $K$ neurons on each layer.

  \begin{equation}
    TKNC(D_{test},K) =  \frac{|\bigcup_{x^i \in D_{test}} (\bigcup_{1 \leq j \leq A} top^j_K(x^i)))|} {N}
  \end{equation}

 \subsubsection{k-multi-section neuron coverage}
 
 Given a neuron $c$,  the multi-section neuron coverage measures how thoroughly the given set of test instances  covers the range $[L_{c},H_{c}]$. The range is divided into $k>0$ equal sections,  called multi-sections.  A multi-section $S^s_{c}, \; s \in \{1,\dots,k\}$ is said to be covered if $\phi(x^i,c) \in S^s_{c}$ for $x^i \in D_{test}$.

 The  k-multi-section neuron coverage for neuron $c$ is defined \cite{Ma_et_al:2018} as the ratio between the number of sections covered by $D_{test}$ and $k$, 

 \begin{equation}
    \overline{KMN}(c) = \frac{ \left | \{S^s_{c}| \, \exists x^i \in D_{test}: \phi(x^i,c) \in S^s_{c} \} \right|}{k}
 \end{equation}
 
 The k-multi-section neuron coverage of an MLP \cite{Ma_et_al:2018} is defined as:

  \begin{equation}
    KMN(D_{test},k) =  \frac{ \sum_{c}   \overline{KMN}(c) }{k \cdot N }
 \end{equation}

\subsubsection{Neuron boundary coverage and strong neuron activation coverage}
     
  A test input $x^i \in D_{test}$ is said to be located in the corner-case region of an MLP if thre is a neuron $c$ such that $\phi(x^i,c)$ is lower than $L_{c}$  or higher than $H_{c}$. 

  To cover corner-case regions of MLPs, the sets of covered corner-case regions are defined as: 

  \begin{align}
     LCN &=&  \{c| \, \exists x^i \in D_{test}: \phi(x^i,c) \in (-\infty,L_{c}) \}\\
     UCN &=&  \{c| \, \exists x^i \in D_{test}: \phi(x^i,c) \in (H_{c},+\infty) \}
  \end{align}

  The neuron boundary coverage (NBC) measures how many corner-case regions have been covered by the given test input set $D_{test}$. 

  \begin{equation}
    NBC(D_{test}) =  \frac{|LCN|+|UCN|}{2 \cdot N }
  \end{equation}

  The strong neuron activation coverage (SNAC) measures how many corner cases, with respect to the upper boundary value, have been covered by the given test inputs $D_{test}$.
  
  \begin{equation}
    SNAC(D_{test}) = \frac{|UCN|}{N}
  \end{equation}

\section{Related work}   \label{sec:RELATED_WORK}

 \subsection{Neural network verification}
 
 In the literature, the use of neuron coverage metrics is mainly associated with the evaluation and creation of test instances for verification of neural networks \cite{Ma_et_al:2018,Pei_et_al:2017,Yan_et_al:2020}. Test prioritization consists of ranking the raw inputs to a model according to their potential to improve it. A typical form to improve the model is by uncovering  unexpected behavior from the model that could lead to its enhancement.  Usually, the neuron metric of choice is evaluated for each of the test examples that are then sorted in descending order of coverage amount. The top ranking test instances are given higher priority.   

 Pei et al. \cite{Pei_et_al:2017} introduced  the concept of neuron coverage and used it for white-box testing of deep learning systems. They reported that  neuron coverage is a better metric than code coverage for measuring the comprehensiveness of the DNN test inputs, and that inputs from different classes of a classification problem usually activate more unique neurons than inputs that belong to the same class.  Ma et al. \cite{Ma_et_al:2018} extended the set of neuron and layer coverage metrics,  and used them  combined with the creation of adversarial examples, to quantify the defect detection ability of test data on DNNs. In  \cite{Gerasimou_et_al:2020}, neuron importance analysis was introduced as a way to identify neurons that play a more important role for decision-making within the neural network. The authors show that the introduced metric can detect those neurons of convolutional networks that are more sensitive to changes in relevant pixels of a given input.

  Lee et al. \cite{Lee_et_al:2022} proposed the application of neuron coverage metrics for a problem that is not directly related to test selection.  They applied these metrics as the basis for  neuron selection for gradient-based white-box testing of neural networks. These white-box testing methods require the computation of the gradient of neurons to quantify their behavior. Since such a computation can be expensive, some authors propose strategies for selecting or prioritizing neurons. Examples of such strategies  include the random selection of un-activated neurons \cite{Pei_et_al:2017}, or the identification of neurons near the activation threshold \cite{Guo_et_al:2018}. The use of neuron coverage metrics for neuron prioritization adds to the repertoire of existing methods and indicates that neuron coverage metrics can also be used for distinguishing or categorizing different behaviors or roles of the neural network components. 
 
   The effectiveness of coverage-based methods for test prioritization has also been questioned in a number of works where other statistical-based methods for evaluating neural networks were proposed  \cite{Feng_et_al:2020,Weiss_and_Tonella:2022}. Other authors report   \cite{Yang_et_al:2022} that coverage-driven methods are less effective than gradient-based methods for uncovering defects and improving neural network robustness.
   
  A number of works have recently investigated the suitability of neuron coverage metrics to evaluate other machine learning paradigms. For instance, Trujillo et al. \cite{Trujillo_et_al:2020} present a preliminary study on the  use of these metrics for testing deep reinforcement learning (DeepRL) systems. They compute the correlation between coverage evolutionary patterns of the RL process and the rewards. They conclude that neuron coverage is not sufficient to reach substantial conclusions about the design or structure of DeepRL networks.

   None of this previous research has employed the neuron-coverage metric as a way to search in the space of neural architectures or to find the solution of semi-supervised problems.


    \subsection{Neuroevolution for semi-supervised problems}

 
      There are several papers that address semi-supervised learning using evolutionary optimization techniques (e.g., see \cite{Silva_et_al:2018} for a discussion of some of these approaches). We briefly cover only some of the  papers that describe research in this area, with a focus on methods that share some commonality with our contribution. 

      In \cite{Fitzgerald_et_al:2015}, Fitzgerald et al. addressed semi-supervised problems using grammatical evolution (GE). They employ a grammar codifying if-then rules and evolve programs able to assign  instances to different clusters based on their features. Unlabeled instances are used to measure the clustering performance by means of the  silhouette co-efficient or silhouette score (SC) \cite{Rousseeuw:1987}, and the labeled data is used to measure the performance of the model in terms of classification accuracy.  The fitness function is computed as the sum of the aforementioned two scores. While, in this paper, we also propose the  fitness evaluation of each evolved model as a combination of different scores respectively computed using labeled and unlabeled data, the neuron coverage metrics are fundamentally different from SC and other clustering scores. They are not associated to computing the performance on any auxiliary or target task. Furthermore, the evolutionary algorithm that we use to evolve neural networks is a genetic algorithm (GA) working on a list-based representation of neural network architectures. 

      Another approach to semi-supervised classification problems is self-labelling, or retraining, in which the modeled trained on labeled instances is then used to make predictions on unlabeled instances. These predictions are used as pseudo-labels of the unlabeled examples and used for retraining the classifier. A similar approach was presented in  \cite{Kamalov_et_al:2021} where the authors combined the PageRank and PCA algorithms with a variant of genetic programming (GP) specifically tailored for non-linear symbolic regression. The algorithm was tested on three time series datasets and it was reported that the performance of the hybrid-algorithm overcomes the two algorithms individually. There are other papers that propose the application of GP to semisupervised problems, they mainly use a tree-based program representation \cite{Silva_et_al:2018} and also apply variants of self-labeling strategies \cite{Arcanjo_et_al:2011}.

       There is an increasing number of works \cite{Rasmus_et_al:2015,Tarvainen_and_Valpola:2017} that propose semi-supervised learning methods for \emph{fixed neural networks}, i.e., the architecture of the network is not changed as part of the semi-supervised approach. For example, 
       the consistency regularization method introduced in \cite{Rasmus_et_al:2015} evaluates each data point with and without added artificial noise, and then computes the consistency cost between the two predictions. Only recently, the question of semi-supervised classification in NAS has been addressed.
       In \cite{Pauletto_et_al:2022}, two semi-supervised approaches are applied to semantic segmentation. The algorithm jointly optimizes an  architecture of a neural network and its parameters.  This approach works minimizing the weighted sum of a supervised  loss, and two unsupervised losses. As in previous examples discussed in this section, this approach requires the definition of an auxiliary task (e.g., clustering) and the model is evaluated according to its performance on all the tasks.

             
             


  \section{Neuroevolutionary approach}  \label{sec:NE_APPROACH}

 The neuroevolutionary approach we use is based on the application of a GA with genetic operators designed to work on a list representation of tensorflow programs. In this section, we explain the main components of the algorithm, and in the next section we focus on the main contributions of this paper which are related to the way in which the fitness functions are implemented to deal with the semi-supervised learning scenario.

\subsection{Neural network representation}\label{sec:listas}

In this work, the evolved DNNs are standard, feed-forward, sequential MLPs that are characterized by the following aspects:

\begin{itemize}
	\item Number of hidden layers: Since we consider standard feed-forward sequential architectures, a single integer is enough to encode this aspect.
	\item Initialization functions: The weights in any given layer can be initialized in a different manner and this can condition the local optimum the network reaches. It consists of a list of indices of initialization functions.
	\item Activation functions: Similarly, the activation functions applied in each layer is not fixed. Similarly to the initialization functions, it consists of a list of indices. 
	\item Dropout: Also a per-layer characteristic, this is implemented by means of a list of Boolean elements determining whether the application of dropout after the activation functions should be applied.
	\item Batch normalization: Similarly to the previous aspect, this consists of a list of Boolean elements indicating whether each layer implements batch normalization before the activation functions.
\end{itemize}

The evolvable components have a number of options for variation: 

\begin{itemize}
	\item The DNN weights can be initialized by drawing values from a \texttt{normal} or \texttt{uniform} distribution, or by applying the \texttt{xavier} \cite{Glorot_and_Bengio:2010} variation of the \texttt{normal} initialization.
	\item The following activation functions can be applied to the layers of the DNN: \texttt{Identity}, \texttt{ReLU}, \texttt{eLU}, \texttt{Softplus}, \texttt{Softsign}, \texttt{Sigmoid}, \texttt{Hyperbolic Tangent}. 
	
\end{itemize}

 \subsection{A GA with list-based encoding }
 
 Because the DNN's parameters are encoded using lists, we define a list-based DNN descriptor which specifies the network architecture as well as other parameters, such as the loss function, weight initialization functions, etc. This can be considered a declarative representation, as it exclusively contains the specification of the network, the weights being left outside of the evolutionary procedure. Algorithm~\ref{Alg:GA_DNN} shows the pseudocode of the GA.

\begin{algorithm}[ht]
	Set $t\Leftarrow 0$. Create a population $D_0$ by generating $N$ random DNN descriptions\;
	\While{halting condition is not met}{
		Evaluate $D_t$ using the fitness function\;
		From $D_t$, select a population $D_t^S$ of $Q \leq N$ solutions according to a selection method\;
		Apply mutation with probability $p_m=1-p_x$ to $D_t^S$ and create the offspring set $O_t$. Choice of the mutation operator is made uniformly at random\;
		Create $D_{t+1}$ by using the selection method over $\{D_{t},O_{t}\}$\;
		$t \Leftarrow t+1$\;
	}
	\caption{GA for evolving DNN.}
	\label{Alg:GA_DNN}
\end{algorithm}

\subsection{Genetic operators}

The operators used to mutate individuals are the following:

\begin{itemize}
	\item The \textbf{layer\_change} operator randomly reinitializes the description of a layer chosen at random, e.g., its weight initialization and activation functions; and the number of neurons. 
	\item The \textbf{add\_layer} operator introduces a new (randomly initialized) layer in a random position of the DNN.
	\item The \textbf{del\_layer} operator deletes a randomly chosen layer in the DNN.
	\item The \textbf{activ\_change} operator changes the activation function of a random layer to another function, chosen at random.
	\item The \textbf{weight\_change} operator, similarly to \textbf{activ\_change}, changes the function used to obtain the initial weights of the DNN layer.
\end{itemize}

\subsection{Implementation}

We have used the implementation of the neuron coverage metrics\footnote{Available from \url{https://github.com/DeepImportance/deepimportance_code_release}} developed as part of the work presented in \cite{Gerasimou_et_al:2020}. To implement the neuron coverage based neuroevolutionary approach, we used the \texttt{deatf} library \footnote{Available from \url{https://github.com/IvanHCenalmor/deatf}} which is an extension to Tensorflow2 of the Evoflow library  \footnote{Available from \url{https://github.com/unaigarciarena/EvoFlow}} \cite{Garciarena_et_al:2020}, originally conceived to evolve neural networks implemented in tensorflow\cite{Abadi_et_al:2016}.

In these libraries, the representation of the DNN is split into two types of classes: The representation of the DNN architectures is contained in \texttt{NetworkDescriptor}, which encompasses all the lists mentioned in Section~\ref{sec:listas}. The architecture is implemented in a tensorflow DNN, which is contained in the \texttt{Network} class. We use the  \texttt{MLPDescriptor} and  \texttt{MLP} classes conceived to deal with multi-layer perceptrons.

Different selection operators are available in EvoFlow through the \texttt{DEAP} library \cite{Fortin_et_al:2012}. We use the truncation selection strategy.

 \section{Network training and fitness evaluation}   \label{sec:NE_FITNESS_FUNCTION}

  In the application of neuroevolutionary approaches to supervised classification problems, the 
 original dataset $D$ is usually split into three parts $D=D_{train}\cup D_{val} \cup D_{test}$. 
 $D_{train}$ is used to train the network, $D_{val}$  is used to estimate the performance of the trained network by computing the fitness function, and $D_{test}$ is only used at the end of the evolution to assess the quality of the best networks found by the algorithm.  We consider neuroevolutionary scenarios that use this partition of the data. Notice that another validation set could be used for early stopping of the neural network training process. However, we do not consider this type of early stopping strategy. 

  In our approach to semi-supervised problems, we assume that the $D_{val}$ and $D_{test}$ sets will keep all the labels. They will respectively be employed in the usual way  to evaluate the accuracy of the model during the evolution, and at the end of the evolution. $D_{train}$  will have a $q \in [0,1)$ proportion of unlabeled instances and a $1-q$ proportion of labeled instances. The two sets will be respectively named as $D_{train}^l$ and $D_{train}^u$.

 \subsection{Fitness evaluation for the fully-supervised case}

   For the evaluation process, one or more metrics (e.g., accuracy, learning time, etc.) describing the DNN performance could be  computed. For the binary problems addressed in this paper, we use the \emph{balanced accuracy} metric \cite{Brodersen_et_al:2010} that is appropriate to deal with unbalanced classification problems.

    \begin{equation}
     b\_acc(z,D_{val}) = \frac{1}{2}(\frac{TP}{P}+\frac{TN}{N})
    \end{equation}
   where $z$ is the neural network being evaluated, $P$ and $N$ are respectively the number of positive and negative instances in $D_{val}$,  and  $TP$ and $TN$ are respectively the number of correct positive and negative predictions made by the model on instances in $D_{val}$. 
   
   When $q=0$, we have the fully-supervised case in which all the training data are labeled. In this case, for a candidate neural network  $z$, the fitness function $f(z)$ is simply the balanced accuracy $b\_acc(z,D_{val})$.

 \subsection{Fitness evaluation for the semi-supervised case}      

  When $q \in (0,1)$ and $D_{train}^u \neq \emptyset$, the fitness evaluation will also take into account the neural network coverage computed using the unlabeled cases $NNCov(z,D_{train}^u)$,  where $NNCov$ can be one of the following neuron coverage metrics introduced in Section~\ref{sec:COV_METRICS}: $NC$, $TKNC$, $KMN$, $NBC$, and  $SNAC$.

 Finally, the fitness function for the semi-supervised case is defined as:
 
    \begin{equation}
     f(z) = q  \cdot  NNCov(z,D_{train}^u) + (1-q)  \cdot  b\_acc(z,D_{val})
    \end{equation}

 \subsection{Baselines for semi-supervised classification with neuroevolutionary approaches} \label{sec:BASELINES}
 
In order to assess the performance of the introduced algorithms, we implemented two methods inspired by semi-supervised approaches used in the field. 

The first baseline is based on the use of \emph{uncertainty quantifiers}  \cite{Weiss_and_Tonella:2022} which are intended to measure the uncertainty of the model at the time of predicting the class of a given instance. For instance prioritization, examples that are classified  by the model with low confidence (i.e., for binary classification problems, prediction probability close to $0.5$) are of particular interest since they can represent problematic cases. In \cite{Weiss_and_Tonella:2022}, it is argued that simple metrics can be more effective than neuron coverage metrics for test instance prioritization. Therefore, it is a relevant question to determine how these metrics perform in the context of neuroevolutionary optimization.

For evaluating network architectures, we assume that a neural network is more promising when it predicts,  with high certainty, the class of the unlabeled instances. We define the $CERT$ metric as:

 \begin{equation}
     CERT(z,D_{test}) = \frac{\sum{x^i \in D_{test} }max(p(x^i),1-p(x^i))}{|D_{test}|}
 \end{equation}
where $p(x^i)$ is the probability of $x^i$ belonging to class $1$ as assigned by the neural network $z$. 

 This metric is not expensive to compute since it only requires calculating the predictions of the model for all instances in $D^u_{train}$.
 
 The second baseline uses retraining, an approach frequently described in the literature for semi-supervised learning \cite{Arcanjo_et_al:2011,Kamalov_et_al:2021}. Retraining consists of using the model learned on labeled instances to make predictions on the unlabeled instances. The pseudo-labels predicted by the model are then used to retrain it. In some cases, a confidence value is required in order to consider a pseudo-label to be valid for retraining. 
 
 In our baseline, we set a threshold of $p(x^i) \leq 0.4$ in order to consider a class-$0$  pseudo-label to be correct. Similarly,  we set $p(x^i) \geq 0.6$ in order to consider a class-$1$  pseudo-label to be correct. The algorithm initially learns the model using $D^l_{train}$, then it makes predictions for instances in $D^u_{train}$. Subsequently, the unlabeled instances for which prediction thresholds are satisfied are selected and combined with instances in  $D^l_{train}$ to retrain the model. Finally, the fitness function is the balanced accuracy of the retrained model as computed on $D_{val}$. If none of the unlabeled instances satisfy the constraints on the predicted probabilities to be selected, then no retraining is carried out and the fitness value corresponding to the network architecture is the balanced accuracy for $D_{val}$ produced by the network trained on $D^l_{train}$.
 
 The retraining approach which we denote as $RET$  can be computationally costly because it requires conducting the learning process twice. 


\section{Experiments}  \label{sec:EXPE}

 The general aim of the experiments is to determine whether, and to what extent, the use of neuron coverage metrics influences the evolutionary search of  neural network architectures  for semi-supervised classification problems. In particular, we address the following questions:

 \begin{itemize}
     \item How the quality of the evolved classifiers degrade with respect to the amount of unlabeled instances?
     \item Which of the investigated coverage metrics produces a more beneficial effect in the performance of the evolved architectures?
     \item How do the neuroevolutionary approaches based on neuron coverage compare to approaches based on uncertainty quantification and retraining?
 \end{itemize}
 
 We first present the experimental design, including the characteristics of the benchmark. Subsequently, we address the aforementioned questions, presenting the numerical results of the experiments. 

\subsection{Experimental design}

 To create the classification problem benchmark, we follow the approach described in \cite{Fitzgerald_et_al:2015},  in which fully labeled datasets are used and partial labeling is simulated by only considering a random subset of training data to be labeled; the rest of the data set is treated as unlabeled.  Starting from a binary classification problem for which all labels are known, we will simulate the semi-supervised scenario by removing the labels for a proportion $q$ of the instances in the dataset. We will use different values of $q$ to investigate the influence of the amount of missing labels.

  We have selected $20$ binary classification problems included as part of the PMLB library  \cite{Olson_et_al:2017,Romano_et_al:2021}. Each problem has an associated dataset whose characteristics are described in Table~\ref{tab:datasets}. In the table, Imbalance refers to the amount of imbalance in the number of observations corresponding to the two classes. It is calculated by measuring the squared distance of the incidence proportion of each class from perfect balance in the dataset  \cite{Olson_et_al:2017}.

  \begin{table*}
 \begin{center}   
\begin{tabular}{lrrr}
\toprule
                Dataset &  n\_observations &  n\_features &  Imbalance \\
\midrule
       agaricus\_lepiota &            8145 &          22 &       0.00 \\
    analcatdata\_lawsuit &             264 &           4 &       0.73 \\
             australian &             690 &          14 &       0.01 \\
               backache &             180 &          32 &       0.52 \\
                 biomed &             209 &           8 &       0.08 \\
                 breast &             699 &          10 &       0.10 \\
          breast\_cancer &             286 &           9 &       0.16 \\
breast\_cancer\_W &             569 &          30 &       0.06 \\
               breast\_w &             699 &           9 &       0.10 \\
               buggyCrx &             690 &          15 &       0.01 \\
                   bupa &             345 &           5 &       0.00 \\
                  chess &            3196 &          36 &       0.00 \\
                  churn &            5000 &          20 &       0.51 \\
                  cleve &             303 &          13 &       0.01 \\
               coil2000 &            9822 &          85 &       0.78 \\
                  colic &             368 &          22 &       0.07 \\
               credit\_a &             690 &          15 &       0.01 \\
               credit\_g &            1000 &          20 &       0.16 \\
                    crx &             690 &          15 &       0.01 \\
               diabetes &             768 &           8 &       0.09 \\
\bottomrule
\end{tabular}
\caption{Datasets for the binary classification problems.} \label{tab:datasets}
 \end{center}
 \end{table*}

 Each experiment consists of running a neuroevolutionary algorithm for solving a particular binary classification problem. We use a population size of $20$ individuals and $30$ generations. The neural network architectures are constrained to have a maximum depth of $8$, and the maximum number of neurons in each layer was also set to $8$. The batch-size and the number of epochs were respectively set to $10$ and $50$.
 
 Each possible application of the neuroevolutionary algorithm  is  parameterized by the following parameters:
\begin{enumerate}
    \item Classification problem addressed.
    \item Proportion of unlabeled data in $D_{train}$, $q \in \{0, 0.2, 0.4, 0.6, 0.8\}$.
    \item Type of neuron coverage metric used (only when $q>0$), i.e., $NC$, $TKNC$, $KMN$, $NBC$, and $SNAC$.
\end{enumerate}

 For each possible configuration, we have executed $10$ repetitions of the neuroevolutionary search for a total of $20 \times  10 \times (4 \times 5 + 1 )$ $=4200$ experiments. Notice that, when $q=0$, none of the five coverage metrics is used. 

 For each algorithm, and once the evolution has finished, we retrain the architectures encoded by all individuals in the last generation using the dataset $D_{train} \cup D_{val}$. We then make the predictions for $D_{test}$, and compute the balanced accuracy using these predictions. 


\subsection{Numerical results}

\subsubsection{Initial experiments}

\begin{figure}[htbp]
    \begin{center}
   \includegraphics[width=12.0cm]{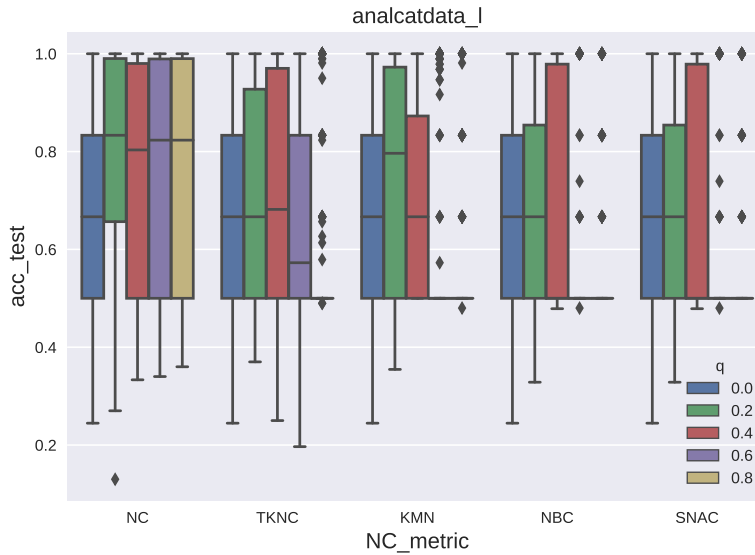}     
   \caption{Results for the analcatdata problem.}
  \label{fig:RESULTS_analcatdata}
 \end{center}
  \end{figure}

\begin{figure}[htbp]
    \begin{center}
   \includegraphics[width=12.0cm]{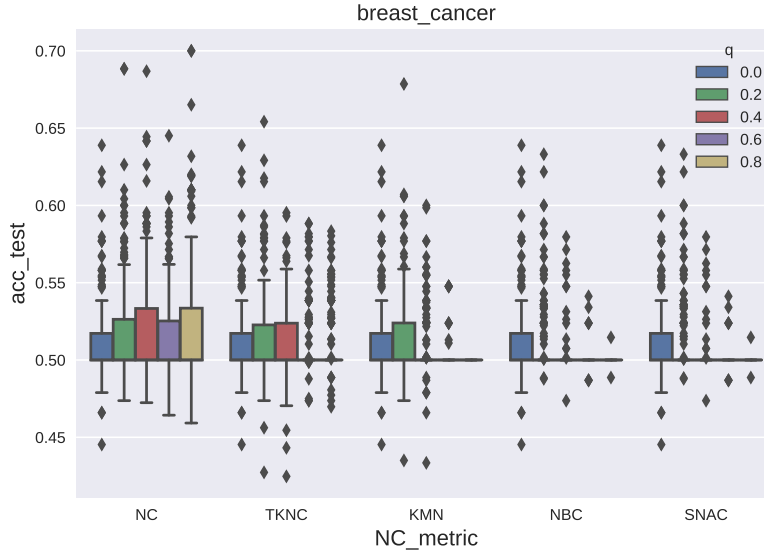}     
   \caption{Results for the breast-cancer problem.}
  \label{fig:RESULTS_breast_cancer}
 \end{center}
  \end{figure}

 For an initial assessment of the influence of $q$ in the performance of the different algorithms, we present in Figure~\ref{fig:RESULTS_analcatdata} and Figure~\ref{fig:RESULTS_breast_cancer} the distribution of the balanced accuracy for two classification problems (analcatdata, and breast\_cancer).
 
 The distribution is computed using the $20$ architectures in the last generation for all the $10$ runs.  In the figures, and to ease the comparison between the algorithms, the results for the fully-supervised problem ($q=0$) are displayed five times, once for each of the neuron coverage metrics. 
 
 The two problems illustrate a different scenario of the performance of the evolved classifiers. The analcatdata problem is easy to solve and, for most of the configurations, the accuracy results are high. This is also an example where the performance of the classifiers does not suffer much when the proportion of unlabeled data is increased. For this problem, among the neuron coverage metrics, NC shows a more stable behavior when $q$ varies.  This example shows that, at least for some problems, using the coverage metrics together with a significant amount of unlabeled data can contribute to obtain a high-performing classifier.

 Figure~\ref{fig:RESULTS_breast_cancer} shows a problem for which the quality of the evolved classifiers is rather poor and all accuracy values are below $0.7$. It is noticeable that, in this problem, the results of the architectures evolved using the $KMN$, $NBC$, and $SNAC$ coverage metrics significantly deteriorate when the amount of unlabeled data is increased. This is a common trend for other problems as will be shown in the following experiments. 

 It is also worth noting that in some cases, such as multiple instances related to the analcatdata database, and instances with $q=20$ and $q=40$ for the breast-cancer problem, the balanced accuracy achieved by the models evolved using neuron-coverage metrics was higher than that obtained using the whole labeling of the database. 

\subsubsection{Evaluation of the algorithms on all datasets}

\begin{figure}[htbp]
    \begin{center}
   \includegraphics[width=12.0cm]{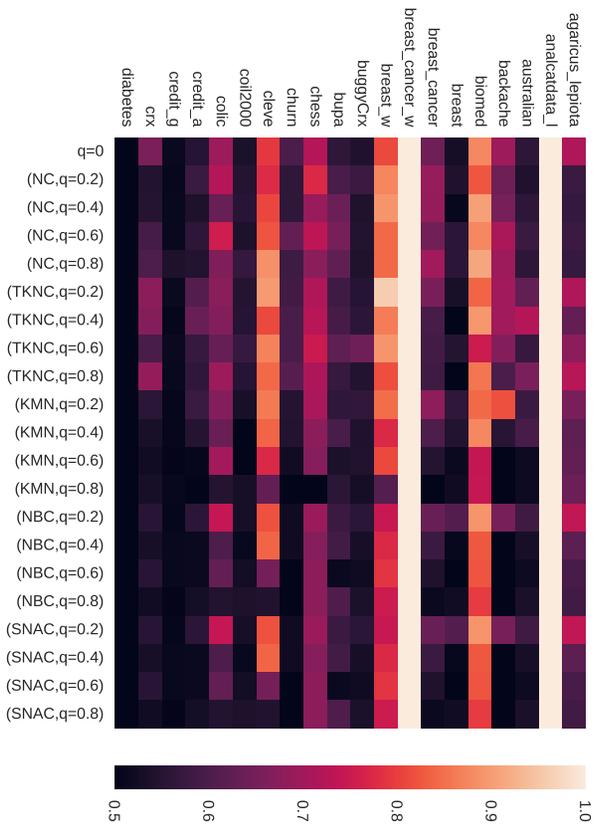}    
   \caption{Balanced accuracy of the best solutions computed for all datasets and metrics.}
  \label{fig:RESULTS_general}
 \end{center}
  \end{figure}

\begin{figure}[htbp]
    \begin{center}
   \includegraphics[width=12.0cm]{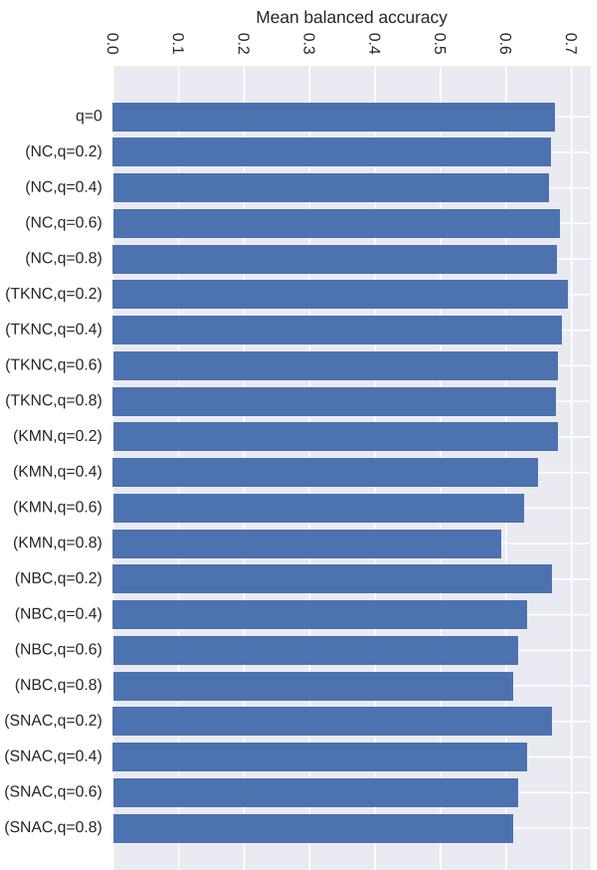}     
   \caption{Mean accuracy of the best classifiers computed for all problems.}
  \label{fig:RESULTS_summ_general}
 \end{center}
  \end{figure}

 Figure~\ref{fig:RESULTS_general} shows the accuracy of the best classifier, in terms of balanced accuracy on the test set, found for each dataset and configuration. The analysis of the figure reveals that, in terms of the best solution found, the amount of unlabeled data does not seem to have a critical impact in terms of the accuracy of the best classifier. However, differences are difficult to spot due to the variability of the problem difficulty among the datasets. Therefore,  we summarize the information contained in  Figure~\ref{fig:RESULTS_general} by computing the average of the balanced accuracy considering the $20$ problems.  This information is shown in Figure~\ref{fig:RESULTS_summ_general}.
 
 As can be seen in Figure~\ref{fig:RESULTS_summ_general}, for three of the metrics, there is a noticeable impact in the accuracy when less labeled data is used for training the network.  For metrics $KMN$, $NBC$, and $SNAC$, as $q$ increases, the mean accuracy decreases. Neuron coverage metrics $NC$ and $TKNC$ confirm  to be more stable and show their capacity to guide the search towards classifiers that are at least as good as those learned using the full set of labeled data. 

 \subsection{Comparison with the baseline algorithms}
 
    In this section, we compare the performance of the introduced algorithms with two methods that were introduced in Section~\ref{sec:BASELINES}. We focus our comparison on the algorithms that use $NC$ and $TKNC$, since they produced the best results in the previous experiments. In the comparison, all the neural network and evolutionary algorithm settings are the same as in previous experiments. The only difference among the algorithms is the way in which the fitness function is computed.

    \begin{table*}
 \tiny
\begin{tabular}{|r|r|rrrr|rrrr|rrrr|rrrr|} \hline
 &  & \multicolumn{4}{c|}{NC}  & \multicolumn{4}{c|}{TKNC} & \multicolumn{4}{c|}{CERT} & \multicolumn{4}{c|}{RET}  \\ \hline
 Ind. DB / q & 0 & 0.20 & 0.40 & 0.60 & 0.80 & 0.20 & 0.40 & 0.60 & 0.80 & 0.20 & 0.40 & 0.60 & 0.80 & 0.20 & 0.41 & 0.60 & 0.80 \\ \hline
  australian & 0.56 & 0.54 & 0.56 & 0.58 & 0.56 & \underline{0.63} & \underline{0.72} & 0.57 & \underline{0.65} & 0.57 & 0.62 & 0.54 & 0.54 & 0.57 & 0.57 & \underline{0.73} & 0.57 \\
  backache & 0.70 & 0.64 & 0.65 & \underline{0.71} & \underline{0.70} & \underline{0.70} & \underline{0.70} & 0.67 & 0.60 & 0.59 & 0.66 & 0.60 & 0.65 & 0.67 & 0.59 & 0.58 & 0.67 \\
  biomed & 0.88 & 0.82 & \underline{0.90} & \underline{0.88} & \underline{0.91} & 0.84 & 0.89 & 0.75 & 0.86 & 0.80 & 0.70 & 0.75 & 0.75 & \underline{0.85} & 0.88 & 0.80 & 0.86 \\
  breast & 0.53 & \underline{0.54} & \underline{0.51} & \underline{0.56} & \underline{0.56} & 0.53 & 0.50 & 0.55 & 0.50 & 0.51 & 0.50 & 0.50 & 0.51 & 0.51 & \underline{0.51} & 0.51 & 0.52 \\
  breast\_cancer & 0.64 & \underline{0.69} & \underline{0.69} & 0.65 & \underline{0.70} & 0.65 & 0.60 & 0.59 & 0.58 & 0.67 & 0.63 & 0.60 & 0.55 & \underline{0.69} & 0.60 & \underline{0.68} & 0.68 \\
  breast\_w & 0.81 & 0.87 & \underline{0.89} & 0.84 & 0.84 & 0.\underline{96} & 0.86 & \underline{0.89} & 0.82 & 0.85 & 0.78 & 0.78 & 0.70 & 0.81 & 0.87 & 0.86 & \underline{0.90} \\
 buggyCrx & 0.54 & \underline{0.58} & 0.54 & 0.55 & 0.54 & 0.55 & \underline{0.56} & \underline{0.64} & 0.55 & 0.55 & 0.52 & 0.52 & 0.52 & 0.54 & 0.55 & 0.54 & \underline{0.56} \\
 bupa & 0.56 & 0.60 & \underline{0.64} & 0.65 & 0.62 & 0.58 & 0.59 & \underline{0.62} & 0.57 & 0.59 & 0.57 & 0.56 & 0.57 & \underline{0.61} & 0.61 & 0.59 & 0.60 \\
  chess & 0.72 & \underline{0.77} & 0.69 & 0.73 & 0.67 & 0.72 & \underline{0.73} & \underline{0.75} & \underline{0.71} & 0.70 & 0.70 & 0.64 & 0.66 & 0.74 & 0.70 & 0.71 & \underline{0.71} \\
   churn & 0.60 & 0.56 & 0.56 & \underline{0.63} & 0.58 & \underline{0.59} & \underline{0.60} & 0.60 & \underline{0.61} & 0.56 & 0.59 & 0.58 & 0.59 & 0.54 & 0.59 & 0.54 & 0.56 \\
\hline
\end{tabular}
    \caption{Comparison of the neuroevolutionary variants based on neuron coverage metrics with the baselines that use the CERT metric and retraining (RET).}
  \label{fig:RESULTS_base_lines}
\end{table*}

 Table~\ref{fig:RESULTS_base_lines} summarizes the results of the comparison among the algorithms for $10$ of the $20$ problems (due to page limit constraints).
 For datasets agaricus\_lepiota and coil2000 the retraining variant had not finished after 15 hours of computation. On the other hand, datasets analcatdat\_l and breast\_cancer\_wand were not included in the analysis since, for all configurations and runs, classifiers with perfect accuracy on the test data were found. 
 
 For each data set, algorithm,  and proportion of unlabeled data, we compute the average accuracy of the best solution in the last population for the $10$ experiments. In  Table~\ref{fig:RESULTS_base_lines}, the algorithm that produces the best result for each dataset and $q$ is underlined. In can be clearly seen that the best results are achieved by using the NC and TKNC neuron coverage metrics. Uncertainty quantification proves to not be a competitive approach for guiding neuroevolution. Retraining can achieve better results than CERT, but at a higher computational cost. 
 
\begin{figure}[htbp]
    \begin{center}
   \includegraphics[width=12.0cm]{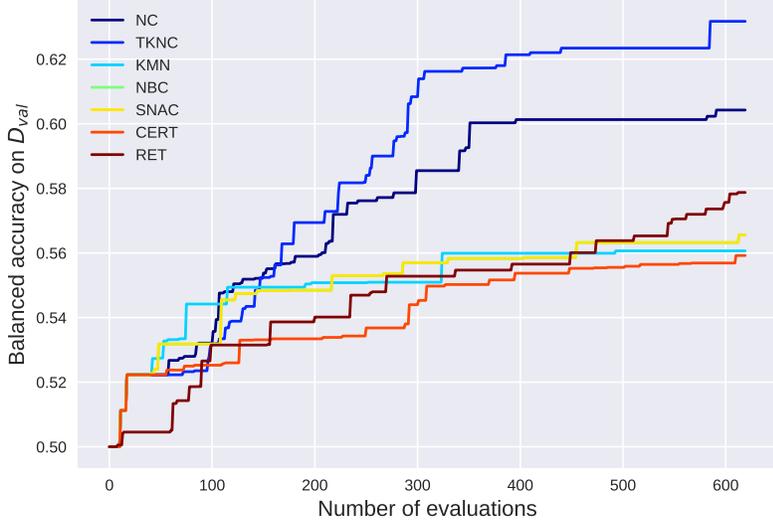}     
   \caption{Evolution of the accuracy on the validation data for the chess dataset and $q=0.8$.}
  \label{fig:RESULTS_EVOL_ACC_80}
 \end{center}
  \end{figure}

  
  We also investigated the dynamics of the neuroevolutionary algorithms during the evolution. Figure~  \ref{fig:RESULTS_EVOL_ACC_80} shows an example of the evolution of the  accuracy of the best solution (computed on validation data $D_{val}$) as the number of evaluations increase from evaluation 1 to evaluation $620$. 
  The illustrative example shown in Figure~\ref{fig:RESULTS_EVOL_ACC_80} corresponds to the chess problem and when $q=0.8$. The balanced accuracy values have been computed as the average of the $10$ runs. 
  
  The curves in Figure~\ref{fig:RESULTS_EVOL_ACC_80} show that all the fitness functions are able to guide the search to areas with better performing neural architectures. The difference in the dynamics is given by the quality of the solutions that are found and the speed of convergence. For the chess problem the neuroevolutionary algorithms using NC and TKNC converge faster to architectures of better accuracy.   Notice that  maximizing the accuracy of the model for $D_{val}$ is not a guarantee of architectures that will generalize to other data (e.g.,  $D_{test}$) but the example illustrates that, in addition to  producing better architectures as shown in Table~ \ref{fig:RESULTS_base_lines}, convergence can be faster.  
  

\section{Conclusions} \label{sec:CONCLU}

   Semi-supervised problems for which labeled instances are difficult or costly to obtain are common in many fields. When neuroevolutionary approaches are applied to these problems the question of how to use the unlabeled data to improve the search for classifiers arises. In this paper, we have proposed the use of neuron coverage metrics as a way to asses how promising each candidate architecture is.  The implicit assumption is that architectures that are better covered by the unlabeled examples are more promising. We have evaluated five different neuron coverage metrics and identified the NC and TKNC metrics as the more stable in terms of the degradation of the results when the number of labeled instances is diminished. Our results also show that for some problems the use of these metrics can even improve the performance of neuroevolutionary search.

\subsection{Future work}

 There are a number of ways in which the work presented in this paper could be extended to deal with other classes of semi-supervised problems. One needed step is to go beyond binary-classification problems to address multi-class problems. The fitness functions proposed in this paper could also be used for the multi-class problems.  Another research direction is defining strategies to deal with multi-label problems learning with incomplete class assignments \cite{Bucak_et_al:2011}. While the computation of the coverage metrics do not change for these problems, the fitness functions should be modified to account for the existence of multiple classification problems. An analysis of the evolved architectures should also be conducted to determine whether the use of the neuron coverage metrics introduces any bias in the type of neural network components (e.g., the type of activation functions) that are included in the best solutions.

 While we have focused on semi-supervised problems, neuron coverage metrics could be used in other scenarios where neuroevolution is applied. They could be applied as an additional regularization mechanism that prioritizes architectures that are fully covered by the inputs of the problem. They could be employed, in single or multi-objective scenarios,  as a diversification mechanism for problems where a large number of candidate neural architectures have the same value of the objective function being optimized. Finally, neuroevolution based on neuron coverage metrics could be used combined with adaptive instance selection for early verification (and correction) of neural networks to be deployed in machine learning systems.

\bibliographystyle{unsrt}
\bibliography{Thesbib_split1,Thesbib_split2} 

\begin{thebibliography}{10}

\bibitem{Galvan_and_Mooney:2021}
Edgar Galv{\'a}n and Peter Mooney.
\newblock Neuroevolution in deep neural networks: Current trends and future
  challenges.
\newblock {\em IEEE Transactions on Artificial Intelligence}, 2(6):476--493,
  2021.

\bibitem{Stanley_et_al:2019}
Kenneth~O Stanley, Jeff Clune, Joel Lehman, and Risto Miikkulainen.
\newblock Designing neural networks through neuroevolution.
\newblock {\em Nature Machine Intelligence}, 1(1):24--35, 2019.

\bibitem{Blum_and_Mitchell:1998}
Avrim Blum and Tom Mitchell.
\newblock Combining labeled and unlabeled data with co-training.
\newblock In {\em Proceedings of the Eleventh Annual Conference on
  Computational Learning Theory}, pages 92--100. ACM, 1998.

\bibitem{Blum_and_Chawla:2001}
Avrim Blum and Shuchi Chawla.
\newblock Learning from labeled and unlabeled data using graph mincuts.
\newblock In {\em Proceedings of the Eighteenth International Conference on
  Machine Learning}, ICML-2001, page 19–26, San Francisco, CA, USA, 2001.
  Morgan Kaufmann Publishers Inc.

\bibitem{Bennett_and_Demiriz:1998}
Kristin Bennett and Ayhan Demiriz.
\newblock Semi-supervised support vector machines.
\newblock {\em Advances in Neural Information Processing Systems}, 11, 1998.

\bibitem{Fitzgerald_et_al:2015}
Jeannie~M Fitzgerald, R~Muhammad~Atif Azad, and Conor Ryan.
\newblock Geml: Evolutionary unsupervised and semi-supervised learning of
  multi-class classification with grammatical evolution.
\newblock In {\em 2015 7th International Joint Conference on Computational
  Intelligence (IJCCI)}, volume~1, pages 83--94. IEEE, 2015.

\bibitem{Kamalov_et_al:2021}
Mikhail Kamalov, Aur{\'e}lie Boisbunon, Carlo Fanara, Ingrid Grenet, and
  Jonathan Daeden.
\newblock Pazoe: classifying time series with few labels.
\newblock In {\em 2021 29th European Signal Processing Conference (EUSIPCO)},
  pages 1561--1565. IEEE, 2021.

\bibitem{Silva_et_al:2018}
Sara Silva, Leonardo Vanneschi, Ana~IR Cabral, and Maria~J Vasconcelos.
\newblock A semi-supervised {G}enetic {P}rogramming method for dealing with
  noisy labels and hidden overfitting.
\newblock {\em Swarm and Evolutionary Computation}, 39:323--338, 2018.

\bibitem{Liu_et_al:2020c}
Chenxi Liu, Piotr Doll{\'a}r, Kaiming He, Ross Girshick, Alan Yuille, and
  Saining Xie.
\newblock Are labels necessary for neural architecture search?
\newblock In {\em Computer Vision--ECCV 2020: 16th European Conference,
  Glasgow, UK, August 23--28, 2020, Proceedings, Part IV 16}, pages 798--813.
  Springer, 2020.

\bibitem{Pauletto_et_al:2022}
Lo{\"\i}c Pauletto, Massih-Reza Amini, and Nicolas Winckler.
\newblock Se2nas: Self-semi-supervised architecture optimization for semantic
  segmentation.
\newblock In {\em 2022 26th International Conference on Pattern Recognition
  (ICPR)}, pages 54--60. IEEE, 2022.

\bibitem{Ma_et_al:2018}
Lei Ma, Felix Juefei-Xu, Fuyuan Zhang, Jiyuan Sun, Minhui Xue, Bo~Li, Chunyang
  Chen, Ting Su, Li~Li, Yang Liu, et~al.
\newblock Deepgauge: Multi-granularity testing criteria for deep learning
  systems.
\newblock In {\em Proceedings of the 33rd ACM/IEEE international conference on
  automated software engineering}, pages 120--131, 2018.

\bibitem{Pei_et_al:2017}
Kexin Pei, Yinzhi Cao, Junfeng Yang, and Suman Jana.
\newblock Deepxplore: Automated whitebox testing of deep learning systems.
\newblock In {\em Proceedings of the 26th Symposium on Operating Systems
  Principles}, pages 1--18, 2017.

\bibitem{Yang_et_al:2022}
Zhou Yang, Jieke Shi, Muhammad~Hilmi Asyrofi, and David Lo.
\newblock Revisiting neuron coverage metrics and quality of deep neural
  networks.
\newblock In {\em 2022 IEEE International Conference on Software Analysis,
  Evolution and Reengineering (SANER)}, pages 408--419. IEEE, 2022.

\bibitem{Lee_et_al:2022}
Seokhyun Lee, Sooyoung Cha, Dain Lee, and Hakjoo Oh.
\newblock Effective white-box testing of deep neural networks with adaptive
  neuron-selection strategy.
\newblock In {\em Proceedings of the 29th ACM SIGSOFT International Symposium
  on Software Testing and Analysis}, pages 165--176, 2020.

\bibitem{Gerasimou_et_al:2020}
Simos Gerasimou, Hasan~Ferit Eniser, Alper Sen, and Alper Cakan.
\newblock Importance-driven deep learning system testing.
\newblock In {\em Proceedings of the ACM/IEEE 42nd International Conference on
  Software Engineering}, pages 702--713, 2020.

\bibitem{Hernandez_et_al:2016}
Jer{\'o}nimo Hern{\'a}ndez-Gonz{\'a}lez, Inaki Inza, and Jose~A Lozano.
\newblock Weak supervision and other non-standard classification problems: a
  taxonomy.
\newblock {\em Pattern Recognition Letters}, 69:49--55, 2016.
\newblock Surveys/Weak supervision and other non-standard classification
  problems: A taxonomy.pdf.

\bibitem{Yan_et_al:2020}
Shenao Yan, Guanhong Tao, Xuwei Liu, Juan Zhai, Shiqing Ma, Lei Xu, and Xiangyu
  Zhang.
\newblock Correlations between deep neural network model coverage criteria and
  model quality.
\newblock In {\em Proceedings of the 28th ACM Joint Meeting on European
  Software Engineering Conference and Symposium on the Foundations of Software
  Engineering}, pages 775--787, 2020.

\bibitem{Guo_et_al:2018}
Jianmin Guo, Yu~Jiang, Yue Zhao, Quan Chen, and Jiaguang Sun.
\newblock {Dlfuzz}: Differential fuzzing testing of deep learning systems.
\newblock In {\em Proceedings of the 2018 26th ACM Joint Meeting on European
  Software Engineering Conference and Symposium on the Foundations of Software
  Engineering}, pages 739--743, 2018.

\bibitem{Feng_et_al:2020}
Yang Feng, Qingkai Shi, Xinyu Gao, Jun Wan, Chunrong Fang, and Zhenyu Chen.
\newblock Deepgini: prioritizing massive tests to enhance the robustness of
  deep neural networks.
\newblock In {\em Proceedings of the 29th ACM SIGSOFT International Symposium
  on Software Testing and Analysis}, pages 177--188, 2020.

\bibitem{Weiss_and_Tonella:2022}
Michael Weiss and Paolo Tonella.
\newblock Simple techniques work surprisingly well for neural network test
  prioritization and active learning (replicability study).
\newblock In {\em Proceedings of the 31st ACM SIGSOFT International Symposium
  on Software Testing and Analysis}, pages 139--150, 2022.

\bibitem{Trujillo_et_al:2020}
Miller Trujillo, Mario Linares-V{\'a}squez, Camilo Escobar-Vel{\'a}squez, Ivana
  Dusparic, and Nicol{\'a}s Cardozo.
\newblock Does neuron coverage matter for deep reinforcement learning? a
  preliminary study.
\newblock In {\em Proceedings of the IEEE/ACM 42nd International Conference on
  Software Engineering Workshops}, pages 215--220, 2020.

\bibitem{Rousseeuw:1987}
Peter~J Rousseeuw.
\newblock Silhouettes: a graphical aid to the interpretation and validation of
  cluster analysis.
\newblock {\em Journal of Computational and Applied Mathematics}, 20:53--65,
  1987.

\bibitem{Arcanjo_et_al:2011}
Filipe de~Lima Arcanjo, Gisele~Lobo Pappa, Paulo~Viana Bicalho, Wagner
  Meira~Jr, and Altigran~Soares da~Silva.
\newblock Semi-supervised genetic programming for classification.
\newblock In {\em Proceedings of the 13th annual Conference on Genetic and
  Evolutionary Computation}, pages 1259--1266, 2011.

\bibitem{Rasmus_et_al:2015}
Antti Rasmus, Mathias Berglund, Mikko Honkala, Harri Valpola, and Tapani Raiko.
\newblock Semi-supervised learning with ladder networks.
\newblock {\em Advances in Neural Information Processing Systems}, 28, 2015.

\bibitem{Tarvainen_and_Valpola:2017}
Antti Tarvainen and Harri Valpola.
\newblock Mean teachers are better role models: {W}eight-averaged consistency
  targets improve semi-supervised deep learning results.
\newblock {\em Advances in Neural Information Processing Systems}, 30, 2017.

\bibitem{Glorot_and_Bengio:2010}
Xavier Glorot and Yoshua Bengio.
\newblock Understanding the difficulty of training deep feedforward neural
  networks.
\newblock In {\em Proceedings of the Thirteenth International Conference on
  Artificial Intelligence and Statistics}, pages 249--256, 2010.

\bibitem{Garciarena_et_al:2020}
Unai Garciarena, Roberto Santana, and Alexander Mendiburu.
\newblock {EvoFlow}: A {P}ython library for evolving deep neural network
  architectures in tensorflow.
\newblock In {\em Proceedings of the 2020 IEEE Symposium Series on
  Computational Intelligence (SSCI-2020)}, pages 2288--2295. IEEE, 2018.

\bibitem{Abadi_et_al:2016}
Mart{\'\i}n Abadi, Paul Barham, Jianmin Chen, Zhifeng Chen, Andy Davis, Jeffrey
  Dean, Matthieu Devin, Sanjay Ghemawat, Geoffrey Irving, Michael Isard, et~al.
\newblock Tensorflow: a system for large-scale machine learning.
\newblock In {\em Osdi}, volume~16, pages 265--283. Savannah, GA, USA, 2016.

\bibitem{Fortin_et_al:2012}
F{\'e}lix-Antoine Fortin, De~Rainville, Marc-Andr{\'e}~Gardner Gardner, Marc
  Parizeau, Christian Gagn{\'e}, et~al.
\newblock {DEAP}: Evolutionary algorithms made easy.
\newblock {\em The Journal of Machine Learning Research}, 13(1):2171--2175,
  2012.

\bibitem{Brodersen_et_al:2010}
Kay~Henning Brodersen, Cheng~Soon Ong, Klaas~Enno Stephan, and Joachim~M
  Buhmann.
\newblock The balanced accuracy and its posterior distribution.
\newblock In {\em 20th International Conference on Pattern Recognition}, pages
  3121--3124. IEEE, 2010.

\bibitem{Olson_et_al:2017}
Randal~S. Olson, William La~Cava, Patryk Orzechowski, Ryan~J. Urbanowicz, and
  Jason~H. Moore.
\newblock {PMLB}: a large benchmark suite for machine learning evaluation and
  comparison.
\newblock {\em BioData Mining}, 10(36):1--13, Dec 2017.

\bibitem{Romano_et_al:2021}
Joseph~D Romano, Trang~T Le, William La~Cava, John~T Gregg, Daniel~J Goldberg,
  Praneel Chakraborty, Natasha~L Ray, Daniel Himmelstein, Weixuan Fu, and
  Jason~H Moore.
\newblock {PMLB v1.0}: an open source dataset collection for benchmarking
  machine learning methods.
\newblock {\em arXiv preprint arXiv:2012.00058v2}, 2021.

\bibitem{Bucak_et_al:2011}
Serhat~Selcuk Bucak, Rong Jin, and Anil~K Jain.
\newblock Multi-label learning with incomplete class assignments.
\newblock In {\em CVPR 2011}, pages 2801--2808. IEEE, 2011.

\end{thebibliography}

\end{document}